\documentclass{article}

\usepackage{PRIMEarxiv}

\usepackage{graphicx}
\usepackage{algorithm}
\usepackage{algorithmic}
\usepackage{hyperref}       
\usepackage{url}
\usepackage[utf8]{inputenc} 
\usepackage[T1]{fontenc}    
\usepackage{hyperref}       
\usepackage{url}            
\usepackage{booktabs}       
\usepackage{amsfonts}       
\usepackage{nicefrac}       
\usepackage{microtype}      
\usepackage{lipsum}
\usepackage{fancyhdr}       
\usepackage{graphicx}  
\usepackage{amsmath,amssymb,amsfonts}
\graphicspath{{media/}}     

\title{Ada-QPacknet - adaptive pruning with bit width reduction as an efficient continual learning method without forgetting
\thanks{\textit{\underline{}}
\textbf{Paper accepted at ECAI 2023}} 
}

\author{
  Marcin Pietron \\
  AGH-UST \\
  Kraków\\
  \texttt{pietron@agh.edu.pl} \\
   \And
  Dominik Zurek \\
  AGH-UST \\
  Kraków\\
  \texttt{dzurek@agh.edu.pl} \\
  \And
  Kamil Faber \\
  AGH-UST \\
  Kraków \\
  \texttt{kfaber@agh.edu.pl} \\
  \And
  Roberto Corizzo \\
  American University \\
  Washington \\
  \texttt{corizzo@american.edu} \\
}

\begin{document}
\maketitle

\begin{abstract}

Continual Learning (CL) is a process in which there is still huge gap between human and deep learning model efficiency. Recently, many CL algorithms were designed. Most of them have many problems with learning in dynamic and complex environments. 
In this work new architecture based approach Ada-QPacknet is described. 
It incorporates the pruning for extracting the sub-network for each task. The crucial aspect in architecture based CL methods is theirs capacity. In presented method the size of the model is reduced by efficient linear and nonlinear quantisation approach. The method reduces the bit-width of the weights format. The presented results shows that low bit quantisation achieves similar accuracy as floating-point sub-network on a well-know CL scenarios.
To our knowledge it is the first CL strategy which incorporates both compression techniques pruning and quantisation for generating task sub-networks.
The presented algorithm was tested on well-known episode combinations and compared with most popular algorithms. Results show that proposed approach outperforms most of the CL strategies in task and class incremental scenarios. 
\end{abstract}


\section{Introduction}
Continual learning (CL) is an emerging machine learning paradigm that aims at designing new methods and strategies to provide accurate analysis in complex and dynamic real-world environments \cite{parisi2019continual}. As models are challenged with multiple tasks over their lifetime, 
a desired property for CL strategies is to maintain a high performance across all tasks. This paradigm is receiving increasing attention from research communities, which led to a number of works being proposed \cite{BAKER2023274,faber2022active,FABER2023248,parisi2019continual,wortsman2020,zhou2023deep}. 
There are three main types of CL strategies: rehearsal (also known as experience replay), regularization, and architectural. 
Architectural strategies are quite powerful in that they allow to adapt and evolve the model to accommodate new tasks, and recent research shows that such approaches are among the most efficient and effective in CL scenarios \cite{kang2022,mallya2017,rusu2016,wortsman2020}.
Forget-free methods \cite{kang2022,mallya2017} accommodate new tasks by assigning a subset of the number of available weights to each task. 
Their effectiveness strongly depends on their ability to effectively leverage model capacity to accommodate as many tasks as possible. 

One important pitfall of existing pruning-based architectural forget-free methods is that they are limited in their ability to efficiently exploit model sparsity since each layer is pruned with the same constant sparsity level. 
As a result, they do not properly tune the number of weights to be removed while preserving classification accuracy \cite{pietron2020retrain, xu2021}. Another general limitation shared by forget-free methods is that they are prone to a quick saturation of the available model capacity, in that they cannot assign more than one value to each weight and are therefore limited by the available number of weights.

In this paper, we propose Ada-QPacknet, a novel forget-free method for continual image classification that solves these limitations by incorporating adaptive pruning with sparsity-level identification and adaptive non-linear weights quantization. 

Specifically, to effectively exploit model sparsity, our proposed Ada-QPacknet incorporates adaptive pruning with different sparsity levels for each layer, where pruning ratios are chosen via fast lottery ticket search. Moreover, to deal with the issue of quickly saturating model capacity, Ada-QPacknet performs an adaptive quantization stage that separates each weight into multiple components, each using a subset of the available 32 bits. This separation allows us to reuse a single weight for more than one task, leading to reduced use of models' capacity and improvements in terms of model efficiency.
At the same time, by exclusively assigning single components of weights to tasks, we overcome the bias limitation that is typically existent in common forget-free methods, which assign multiple tasks to the same weight value. 

In summary, the contributions of our paper are the following:
\begin{itemize}
\item{An adaptive pruning approach that leverages the sensitivity and weights' importance characterizing different layers in deep learning models. This approach allows us to reduce the available set of model's weights assigned to separate tasks, allowing AdaQ-Packnet to efficiently use a subset of the weights without significant losses in accuracy. 

}
\item{A quantization strategy for continual learning, which allows us to split weights into components, each of which can be assigned to a specific task. This capability allows AdaQ-Packnet to better exploit model efficiency and capacity, leveraging the reduced bit-width representation of the weights.} 

\item{An experimental evaluation with three benchmark CL scenarios, which highlights that our proposed method outperforms state-of-the-art rehearsal, regularization, and architectural CL strategies, both in terms of accuracy and exploitation of model capacity.}
\end{itemize}

The paper is structured as follows. 
Section \ref{sec:background} summarizes related works in CL strategies and pruning methods. 
Section \ref{sec:method} describes our proposed Ada-QPacknet method. Section \ref{sec:results} describes our experimental setup, and presents the results extracted in our experiments. Section \ref{sec:conclusions} wraps up the paper with a summary of the results obtained and outlines relevant directions for future work.

\section{Related works}
\label{sec:background}
Continual learning methods are commonly categorized as rehearsal, regularization, and architectural \cite{parisi2019continual}.  
Rehearsal methods use memory to remember some data from past episodes.  
The simplest solution is to store part of the previous training data and interleave them with new training data (Replay \cite{parisi2019continual}). Other popular rehearsal methods include GEM \cite{lopezpaz2017}, A-GEM \cite{chaudhry2019}, and GDumb \cite{prabhu2020}. Regularization methods usually use constraints on the loss function to strengthen connections for already learned patterns. The most popular methods are Synaptic Intelligence (SI) \cite{zenke2017}, which estimates the importance of weights based on weighted regularization loss and gradient. Alternative approaches are LwF \cite{li2017learning}, and EWC \cite{kirk2017}. The most relevant works for this study belong to the architectural category, which represents a promising direction in recent works \cite{kang2022}. 

The architectural methods concentrate on the topology of the neural model. One of the most known and effective methods is CWRStar \cite{lomonaco2019nicv2} which performs weights copy and re-initializes weights at the last layer to accommodate new tasks. More efficient methods are those based on pruning which deals with the issue of increasing model capacity and removes unimportant weights. Notable examples include PackNet \cite{mallya2017}, which divides the model into independent sub-networks to address different tasks. Such an approach is thought to be forget-free since sub-networks are directly mapped to tasks and are not subject to change weights. 
However, a major limitation is the risk of growing model capacity. This issue has recently encouraged researchers to explore and devise pruning methods, to reduce model capacity and storage requirements.

Pruning in neural networks can be structural and non-structural. Works in \cite{aketi,chen,hu,li,liu,wang,zhuang} concentrate on structural pruning, which involves pruning entire channels in filters by means of different techniques.  Non-structural pruning can be carried out on pretrained models without retraining as in \cite{pietron2020retrain,motaz2020}, which results in increased efficiency, or with retraining, as in lottery tickets search \cite{frankle2019}, movement pruning \cite{movement_pruning2020}, and variational dropout \cite{molchanov}. 
Other efficient pruning methods include genetic-based approaches as in \cite{xu2021}, which adopts constant masks and defines its own crossover and mutation operators, and reinforcement techniques as in  \cite{han2018}. Pruning can be conducted in combination with retraining to avoid performance drops caused by weights' removal \cite{pietron2020retrain}. Two recent continual learning methods adopting pruning  approaches are SupSup \cite{wortsman2020} and Winning Subnetworks (WSN) \cite{kang2022}. SupSup finds the optimal binary mask on a fixed backbone network for each new task. On the other hand, WSN introduced significant improvement over Packnet, as it only considers a subset of weights from the previous task to avoid a biased forward transfer. Another recent trend is that expandable model architectures with compression. The Dynamically Expandable Representation (DER) method \cite{kn:yan} adds a new submodule to the main model each time a new task is presented, while leveraging pruning as an additional mode. Authors in \cite{kn:douillard} present the Dytox method, which is based on Transformer attention blocks. The continual learning model consists of an encoder part - common for all tasks and a decoder, which is task-specific. On the other hand, FOSTER \cite{kn:wang} proposes a boosting approach that expands the network by means of a new residual feature extractor and a linear classifier layer.
Both Dytox \cite{kn:douillard} and FOSTER \cite{kn:wang} incorporate a replay strategy to improve the model's efficiency.

Quantization is one of the most efficient techniques for the compression of deep learning models \cite{motaz2020,han2015learning}. Common strategies include the quantization of all coefficients in a single layer with a specified number of bits to represent the integer and fractional parts \cite{kn:anwar, kn:gysel} based on the range of values of the coefficients set. Another strategy is to represent coefficients and data by integer numbers with an appropriate scaling factor. Many  quantization approaches in the literature adopt linear \cite{motaz2020, han2015learning, kn:gysel} or non-linear approaches including clustering \cite{pietronCANDAR}. Quantization can be performed during model training \cite{han2015learning} or can be run on a pre-trained model \cite{motaz2020, pietronCANDAR}. Recently, several methods for low-bit representation have been designed \cite{kn:park2017, kn:zhang2018, kn:jung2018, kn:mcdonell2018, kn:mishra2018}. Many of them can not be run without some significant degradation in accuracy. One known advantage of quantization is the fact that it facilitates the adoption of deep neural networks in specialized hardware accelerators with limited arithmetic bit-width and memory space \cite{xu2021}.
However, to the best of our knowledge, our paper is the first attempt to apply quantization to CL models.

\section{Ada-QPacknet algorithm}
\label{sec:method}
In this section, we describe our proposed Ada-QPacknet -- a forget-free method that leverages adaptive pruning and adaptive quantization to eliminate catastrophic forgetting, effectively exploit model sparsity, and efficiently use model capacity. 

\begin{figure*}[ht]
    \centering
\includegraphics[scale=0.75]{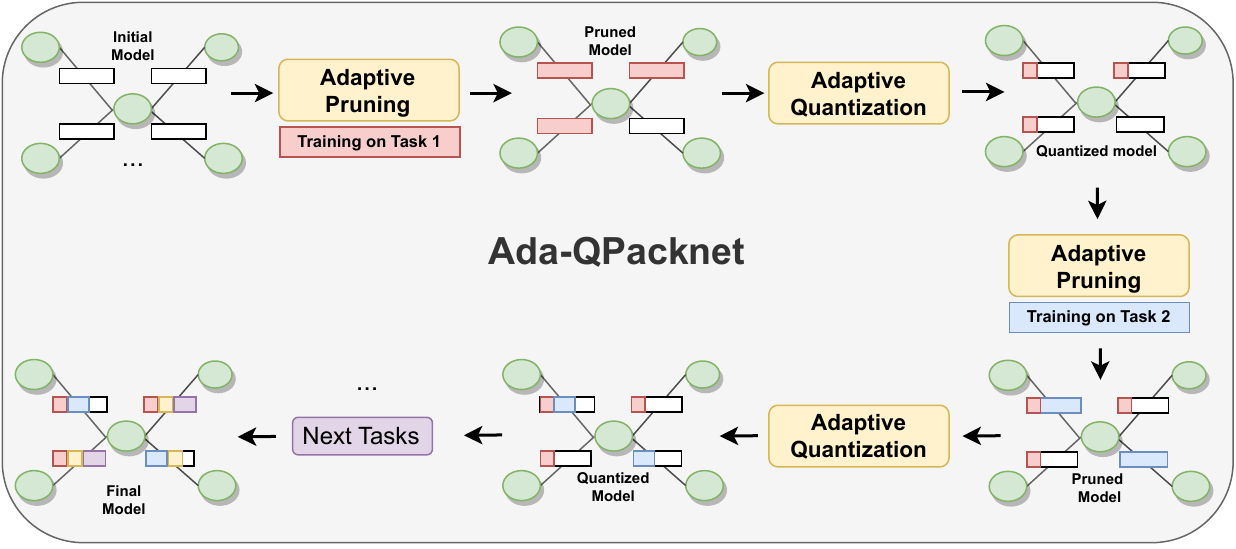}
  \caption{Overview of the proposed Ada-QPacknet method. Initially, the model has a full bit-width representation represented by white rectangles.  The method follows an iterative process that includes adaptive pruning during training and adaptive quantization. Adaptive pruning identifies different sparsity levels for each layer. Adaptive quantization identifies the optimal bit-width for each task (identified by different colors, i.e. red, blue, purple, and yellow) and quantizes weights, improving the exploitation of model capacity. The final model has four weights that support four different tasks, illustrating an efficient use of model sparsity and capacity.}
    \label{fig:method-overview}
\end{figure*}

Figure \ref{fig:method-overview} presents the overview of the Ada-QPacknet algorithm, illustrating the iterative process of adaptive pruning and quantization on a typical training scenario with multiple tasks.

Ada-QPacknet adapts the adaptive pruning process \cite{xu2021, pietron2020retrain} to reduce the size of task sub-networks, since the most significant drawback of state-of-the-art forget-free methods such as PackNet \cite{mallya2017} and WSN \cite{kang2022}, is the constant sparsity level applied to all layers in consecutive tasks. This significantly restricts the adaptability of the method and poses a risk of inefficient use of model capacity.
Another crucial feature of Ada-QPacknet is adaptive quantization, which allows us to assign more than one task to each weight, by dividing its capacity into components. The adaptive nature of the quantization process allows Ada-QPacknet to identify the optimal bit-width for each task, based on the trade-off between the number of bits assigned to each task and the model performance.
In the following subsections, we describe adaptive pruning and adaptive quantization in more detail. 

\subsection{Adaptive Pruning with Sparsity Level Identification}


In this stage, the goal is to perform pruning while identifying separate sparsity levels for each layer. Sparsity levels should be as high as possible to reduce memory occupation leaving more capacity for future tasks. 
The representation of the pruned model $F_{\Theta}^{p}$ 
is the following tuple:
\begin{equation}
F_{\Theta}^{p} = (F_{\Theta}, M),
\label{eq:pruned_model}
\end{equation}
where $F_{\Theta}$ is defined as: 
\begin{equation}
F_{\Theta}(X) = {f_{\theta_L}(f_{\theta_{L-1}}...(f_{\theta_{0}}(X)))}.
\label{eq:model}
\end{equation}
$F_{\Theta}$ is the original model with a set of convolutional and fully-connected layers $f_{\theta_{i}}$, which are defined in the specified order. The $\Theta$ tensor is defined as: 

\begin{equation}
\Theta = \{\theta_{0}, \theta_{1},...,\theta_{L}\},
\label{eq:theta_}
\end{equation}
and contains weights of convolutional and fully-connected layers. $\theta_{i}$ is a tensor of weights between layers $i$ and $i+1$. A single entry $w_{i, k, l}$ represents a weight between $k$-th neuron in layer $i$ and $l$-th neuron in layer $i+1$\footnote{To simplify our notation, we assume that a single weight value is assigned to all tasks. In our method, multiple values for a single weight can be assigned to solve multiple tasks.}.

\begin{figure*}[ht]
   \centering

  \includegraphics[scale=0.55]{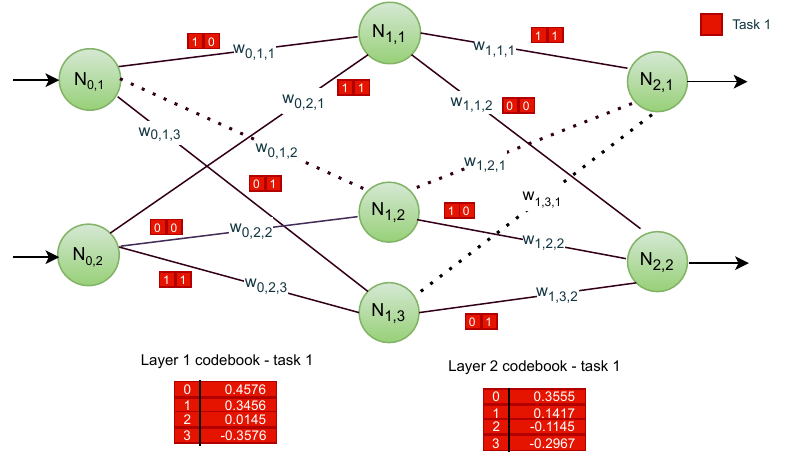}
  \includegraphics[scale=0.55]{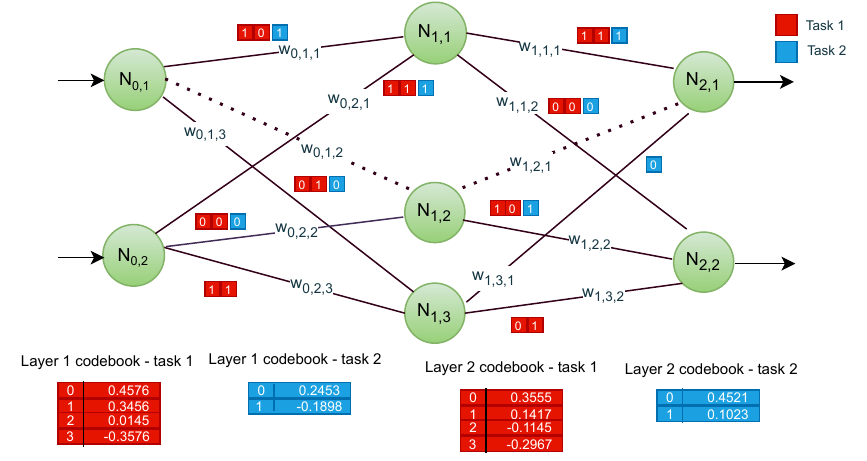}
  \caption{
  The figure presents the status of the model after it is trained, pruned, and quantized for Task 1 (left) and Task 2 (right). During the training stage for Task 1, Adaptive Pruning releases a subset of model weights (dashed lines), which could be used later on to accommodate new tasks. After that, Adaptive Quantization identifies the optimal bit-width and quantizes the representation of the weights. In this example, weight representation of 32-bits is quantized into 2 bits (red squares). Values represented on this bits are codes from a layer-wise codebook and can be translated into full 32-bits weights. This capability allows Ada-QPacknet to divide each weight into multiple components, each of which can be assigned to any of the tasks. The same process is repeated when Task 2 is presented: Adaptive Pruning quantizes the 32-bits weight representation into a single bit (blue squares).}
    \label{fig:algorithm}
\end{figure*}

$M$ is a set of masks: 
\begin{equation}
M = \{M^{0}, M^{1},...,M^{L}\}.
\label{eq:mask_}
\end{equation}
Each mask $M^{i}$ is positioned in between layers $i$ and $i+1$ and has the same shape as $\theta_{i}$.
Masks are used to decide which weights will be used to solve specific tasks. 
A single entry in $M^{i}$ contains a set of numbers indicating in which tasks the weight is used. Therefore, each weight can be assigned to one or more tasks. A graphical representation is shown in Figure \ref{fig:algorithm}.

A single entry in the mask $M^{i}$ is represented as a binary vector: 
\begin{equation}
M^i_{k, l} = \{t_1, t_2, \dots, t_n \},
\label{eq:mask__}
\end{equation}
where $k$ is the source neuron in the layer $i$, $l$ is the destination neuron in layer $i+1$, $t_i$ $\in$ \{0,1\} indicates if a given weight is used by the $i$-th task.


The sparsity level $\Upsilon_i$ is a ratio between the number of weights that are not yet assigned to any tasks and the number of all weights in a mask $M^i$:

\begin{equation}
\Upsilon_i =\frac{|M^{i}|-\sum_{k,l}^{{|M^{i}|}} (|M^{i}_{k, l}| > 0)}{|M^{i}|}
\label{eq:mask___}
\end{equation}



The overall weighted sparsity of the model is defined as: 
\begin{equation}
    \Upsilon = \sum_i^L |\theta_i| \cdot \Upsilon_{i}
    \label{eq:w_sparsity}
\end{equation}




\subsubsection{Generating random lottery ticket mask}
\label{sec:lottery_ticket}
Each time a new task is encountered, for each layer, a number of candidate masks $P_S$ are generated according to the random lottery ticker algorithm, taking into account a sparsity level constraint between $V_{min}$ and $V_{max}$ values. For example, a sparsity level of $0.5$ means that half of all weights will be leveraged to solve the task.
This way, we generate a candidate mask for task $t$ and layer $l$:

\begin{equation}
    CM^{t}_l = \Big\{ e \ | \ \ e \in M^i \wedge \ \ |e| < T_L \Big\}, \  |CM^{t}_l| < |M^i|, \label{eq:random_ticket_mask}
\end{equation}
where $|e|$ is the number of tasks already assigned to a weight $e$, and $T_L$ is a constant that determines the capacity, i.e. the maximum number of tasks that can be assigned to a single weight, according to the desired quantization used. For instance, $T_L=4$ with a weight capacity of 32-bit and a quantization rate is 8-bit. This formulation allows us to sample a subset of available weights from the original mask (according to the sparsity constraint) with zero or more prior assignments to tasks.

The set of candidate masks for all layers is defined as:
\begin{equation}
    CM^t = \{ CM^{t}_0, CM^{t}_1, \dots, CM^{t}_L \}
    \label{eq:candidate_generation}
\end{equation}

Once all candidate masks are generated, the best candidate according to the resulting accuracy for the current layer is selected.

\begin{algorithm}
\begin{algorithmic}[1]
\REQUIRE{$P_{S}$ -- population size}
\REQUIRE{$\Theta$ -- weights of the model}
\REQUIRE{$\alpha$, $\beta$ -- importance of accuracy and sparsity in selection}
\STATE{$\Theta$ $\gets$ initialize the available weights in $\Theta$}
\STATE{$CM^t \gets \emptyset$ // A set of candidate masks}
\STATE{$\Theta^t \gets \emptyset$ // A set of masked weights}
\FOR{$i = 0$ \TO $P_{S}$}
\STATE{$CM^{t,i} \gets $ random lottery ticket mask (eq: \ref{eq:candidate_generation})}


%


\STATE{$\Theta^{t,i}$ = $\Theta$ $\odot$ $CM^{t,i}$} \COMMENT{Masked weights}
\STATE{Execute short training $\Theta^{t,i}$ on task $t$}

\STATE{$CM^t$ $\gets CM^t \cup CM^{t,i}$}
\STATE{$\Theta^t \gets \Theta^t \cup \Theta^{t, i}$}

\ENDFOR \\

\STATE{$A^t \gets \forall_{CM^{t,i} \in CM^t}$ accuracy of  $\Theta \odot$ $CM^{t,i}$ on task $t$} 
\STATE{$\Upsilon^t \gets \forall_{CM^{t,i} \in CM^t}$ compute $\Upsilon_i$ for $CM^{t, i}$ (eq: \ref{eq:w_sparsity})}
\STATE{$id = 
\underset{i \in \{0, \dots, {P_S} \}}{\mathrm{argmax}}\
(\alpha \cdot \frac{A_i}{max(A)}+\beta \cdot \frac{\Upsilon_i}{max(\Upsilon)})$}
\STATE{$M^t$ $\gets$ $CM^{t, idx}$}
\STATE{$\Theta' \gets \Theta^{t, idx}$}
\STATE{Execute full training of $\Theta'$ on task $t$}
\RETURN{$M^t, \Theta' $}
\end{algorithmic}
\caption{Adaptive Pruning for task $t$}
\label{alg:pruning}
\end{algorithm}

Algorithm \ref{alg:pruning} describes the task pruning process for a task $t$. We start with the random initialization of available model weights (line 1). Two empty sets are created: \textit{i)} $CM^t$ used for storing candidate masks (line 2); \textit{ii)} $\Theta^t$ used for storing masked weights (line 3). After that, we generate $P_S$ (population size) candidate masks (lines 4-10). For each generated candidate mask (line 5), we train masked weights (lines 6-7) with a very limited number of epochs. Then, we add the generated candidate mask and trained weights to the respective sets (lines 8-9).
In the next step, we compute the accuracy $A^t$ for all generated candidates (line 11) and weighted sparsity $\Upsilon^t$ (line 12). We identify the best model according to the equation in line 13, where the evaluation simultaneously considers accuracy and sparsity: between two candidates with similar accuracy but dramatically different values of sparsity, the one with the highest level of sparsity is preferred, as we want to preserve as many weights as possible for future tasks.
After this step, we perform the full training stage of the masked weights $\Theta'$ on the current task $t$.

\subsection{Adaptive Quantization}
\label{sec:quantization}
Quantization is a key stage in the Ada-QPacknet algorithm. Its purpose is twofold: \textit{i)} it allows us to assign a single weight to multiple tasks; \textit{ii)} it helps reducing the use of capacity of the model with a specified number of tasks. 
The standard 32-bit single-weight representation is divided into multiple components with a reduced number of bits, and each component can be assigned to  a different task, as shown in Figure \ref{fig:algorithm}. 
For example, one 32-bit single-weight representation may be divided into four 8-bit components that are assigned to any of the tasks.
However, the adaptive nature of this process is not dependent on equal-width components, but allows us to assign a varying number of bits for each component, evaluating the trade-off between bit-width and model performance. For example, a 32-bit single-weight representation may also be divided into 3 components with the following bit-widths: 8, 10, 14, etc.


\begin{algorithm}
\begin{algorithmic}[1]
\REQUIRE{$\psi$ -- Initial bit-width}
\REQUIRE{$Q_q$ -- Model accuracy}
\REQUIRE{$\Theta$ -- Pruned model}
\REQUIRE{$\delta$ -- Maximum loss in accuracy}
\STATE{$\Theta_q, K$ = nonlinear\_quantization($\psi, \Theta$)} \COMMENT{see Algorithm \ref{alg:quant}}
\STATE{$Q_q$ = acc($F$, $\Theta_q)$}
\WHILE{$Q_q$ $<$ $Q$ - $\delta$}
\STATE{$\psi$ = $\psi$ + 1}
\STATE{$\Theta_q, K$ = nonlinear\_quantization($\psi, \Theta$)}
\STATE{$Q_q$ = acc($F$, $\Theta_q)$}
\ENDWHILE
\RETURN{$\psi$ -- optimal bit-width for the task, $\Theta_q$ -- quantized model, $K$ -- codebook}
\caption{Adaptive Quantization for task $t$.}
\label{alg:iter_quant}
\end{algorithmic}
\end{algorithm}
 
Algorithm \ref{alg:iter_quant} presents the details of the adaptive quantization process. It starts with non-linear quantization of the task-pruned model with  minimal values of parameters $\omega$ and $\psi$ (bit-width) (line 1). We leverage the non-linear quantization algorithm detailed in Algorithm \ref{alg:quant}. Then, we check the model accuracy (line 2). Next, if the drop in accuracy is more than the given threshold $\delta$ (line 3) the iterative process of increasing the bit-width is started (lines 3-7) until the drop in performance is deemed to be acceptable.

The non-linear quantization approach devised in this work (see Algorithm \ref{alg:quant}) is inspired by \cite{pietronCANDAR}. 
Specifically, we start with the identification of the number of centroids $\omega$ and the creation of an empty codebook $K$ which will store, for each layer, a mapping between code and weight values (lines 1-2). We also create a copy of the model $\Theta$, where weights will contain codes from the codebook (centroid indices) instead of real values. 
Subsequently, for each layer $i$, we cluster weights values from layer $\theta_i$ leveraging the K-Means clustering algorithm with a specified number of clusters $\omega$ (line 5).
This step returns two outputs: a list of centroids $C$ and the list of weight-to-centroid-index assignment $A$.
We proceed by assigning the corresponding centroid index  to each weight (lines 6-8). 
To keep the mapping between centroid indices and centroids, we build a codebook for each layer (lines 9-11). 
The algorithm returns the copy of the model with quantized weights obtained as codes from the codebook, as well as the codebook required to decipher codes to the real value of the weight.
We note that the number of centroids is directly related to the desired bit-width (as shown in line 1). A low number of centroids implies that the memory needed for storing the weights is decreased. Each weight is stored as an index of the closest centroid. 
A visual representation of the codebook is shown in Figure \ref{fig:algorithm}. 


\begin{algorithm}
\begin{algorithmic}[1]
\REQUIRE{$\psi$ -- desired bit-width}
\REQUIRE{$\Theta$ -- weights of a model}
\STATE{$\omega \gets 2^{\psi}$} \COMMENT{Number of centroids}
\STATE{$K \gets [K_0, K_1, \dots, K_{|\Theta|}]; K_i = \{ \}$} \COMMENT{Empty codebook}
\STATE{$\Theta' \gets$ copy of $\Theta$}
\FOR{$\theta_i$ $\textbf{in}$ $\Theta$}
\STATE{$C, A \gets$ KMeans($\theta_i$, $\omega$)} \COMMENT{$C$: centroids, $A$: weight-to-centroid-index assignment}
\FOR{$\theta_{i, j} \in \theta_i$}
\STATE{$\theta'_{i, j} \gets A[\theta_{i, j}]$} \COMMENT{Use centroid index instead of weight value}
\ENDFOR
\FOR{$c_k \in C$}
\STATE{$K_i[k] \gets c_k$} \COMMENT{Building layer-wise codebook}
\ENDFOR
\ENDFOR
\RETURN{$\Theta', K$}
\caption{Non-linear quantization}
\label{alg:quant}
\end{algorithmic}
\end{algorithm}

\section{Results}
\label{sec:results}
\subsection{Experimental setup}
The experiments were run for three CL scenarios: Permuted MNIST (p-MNIST) \cite{lecun1998mnist}, split CIFAR100 (s-CIFAR100) \cite{Krizhevsky2009LearningML}, and 5 datasets \cite{ebrahimi2020adversarial}, a task-incremental scenario consisting of MNIST, SVHN, FashionMNIST, CIFAR10, not-MNIST, TinyImagenet \cite{le2015tiny}, and Imagenet100 \cite{russakovsky2015imagenet}. The p-MNIST scenario consists of 10 tasks with randomly permuted pixels of the original MNIST (10 tasks with 10 classes each). s-CIFAR100 is divided into 10 tasks with 10 classes each. The 5 datasets scenario is a sequence of 5 tasks with different datasets,  each with 10 classes.   
The TinyImagenet scenario consists of 40 tasks with 5 randomly sampled classes each.
Finally, the Imagenet100 scenario consists of 10 tasks with 10 randomly sampled classes each.

In our experiments, the pruning search population $PS$ is set to 16. Values for  $V_{MIN}$ and $V_{MAX}$ parameters are set to 0.45 and 0.85 for all layers. The $\alpha$ and $\beta$ parameters are set to 0.9 and 0.1.  
The short training during mask search is done with 5 epochs, while the full training for the selected mask uses 50 epochs. For Ada-QPacknet, the learning rate is set resorting to a dynamic scheduler starting from 0.01 and decreasing over time with a minimum of 0.0001 for all datasets except for TinyImagenet (starting from 0.001) and Imagenet100 (constant learning rate of 0.0001).
The adopted models are a two-layer neural network with fully connected layers (p-MNIST), reduced AlexNet (s-CIFAR100) \cite{saha2020gradient}, Resnet-18 (5 datasets and Imagenet), TinyNet (TinyImagenet) in accordance with model backbones used in the WSN paper \cite{kang2022}. 
For weight initialization, we adopt the Xavier initializer. Experiments are executed on a workstation equipped with an NVIDIA A100 GPU. Our experiments involve 5 complete runs for each strategy. Therefore, the total number of executions corresponds to the number of tasks in each scenario multiplied by 5.

As for the metrics, we use the most standard definition of lifelong Accuracy following the description in \cite{diaz2018don}.
Moreover, we also compute the capacity (total memory occupied by each task) inspired by the capacity computation in \cite{kang2022}, defined as follows: 

\begin{equation}
    CAP_t = \sum_{i}^{L}(1-{\Upsilon_{i}})\cdot|\theta_i|\cdot b + |L| \cdot 2^b \cdot (32 + b) \\
    + \sum_{i}^{L}(1-\Upsilon_{i}) \cdot |M^i| 
\end{equation}

The capacity can be regarded as the sum of three components. The first one describes the number of the weights after the pruning multiplied by the bit-width $b$. The second one describes the codebook size. The third one includes all masks' capacity. 

\subsection{Comparison to other methods}

\begin{table*}[htbp]
\small
\centering
\caption{Comparative results in terms of average accuracy for CL strategies in different CL scenarios: Permuted MNIST (p-MNIST), split-CIFAR100 (s-CIFAR100), and 5 datasets.}
\begin{tabular}{ccccc} 
\hline
\textbf{}   & \textbf{p-MNIST} & \textbf{s-CIFAR100} & \textbf{5 datasets} &  \textbf{TinyImagenet} \\ 
\hline

\textbf{Naive}      & 60.12                     & 17.32                  & 33.08  &                         20.27      \\ 
\textbf{CWRStar}    & 31.31                     & 20.84                  & 36,16     &                       24.00       \\ 
\textbf{SI}         & 57.32                     & 19.54                  & 29.42      &                   20.51        \\ 
\textbf{Replay}     & 62.22                     & 19.60                  & 55.24       &                23.14         \\ 
\textbf{CUMULATIVE} & 96.45                     & 36.52                  & 84.44        & 27.53                        \\ 
\hline
\textbf{Packnet}    &  96.31                         & 72,57                  &      92.59   & 55.46 \\ 
\textbf{WSN}        &      96.41                     &       76.38                 &    93.41 &  71.96  \\ 
\hline
\textbf{Ada-QPacknet} &          97.1               &        74.1             &       94.1   & 71.9 \\
\hline
\end{tabular}
\label{table:CL_accuracy}
\end{table*}

\begin{table*}
\small
\centering
\caption{Capacity comparison for pruning-based methods}
\begin{tabular}{ccccc} 
\hline
\textbf{Datasets}   & \textbf{p-MNIST} & \textbf{s-CIFAR100} & \textbf{5 datasets} & \textbf{TinyImagenet} \\ 
\hline
\textbf{Packnet}    &   96.38\%                      &            81.0\%       &         82.86\%   & 188.67\%                  \\ 
\textbf{WSN}        &         77.73\%                  &       99.13\%                &           86.10\%  &   48.65\%                    \\ 
\hline
\textbf{Ada-QPacknet}      &     81.25\%                 &   78.6\%              &   33.7\%            &     112.5\%     \\ 

\hline
\end{tabular}
\label{table:CL_capacity}
\end{table*}

Table \ref{table:CL_accuracy} presents results in terms of average accuracy. In our experimental analysis, we considered representatives from all categories of CL strategies. 
Specifically, we considered  Packnet and WSN (architectural)
, Cumulative and Replay (rehearsal), SI (regularization), and CWRStar (combined architectural and regularization). 
The Naive strategy is a simple strategy where the model is fine-tuned as new tasks are presented, and can be considered as a lower bound. The Cumulative strategy, on the other hand, stores data from all tasks observed so far to retrain the model, and is considered as an upper bound. 

Results highlight that the Ada-QPacknet outperforms all the considered strategies. In p-MNIST, Ada-QPacknet achieves $97.1\%$ ($0.50 \%$ above Cumulative), the WSN method ($96.41\%$) is just behind the Cumulative approach ($96.45\%$). Packnet achieves $96.31\%$, whereas the other methods significantly underperform the architectural strategies (the following method in the ranking is Replay with $62.22\%$). 
In s-CIFAR100, Ada-QPacknet achieves $74.1\%$ average accuracy, achieving second-best performance. The best-performing method is WSN, with a $2.28\%$ increase in performance, although this method presents a higher capacity utilization when compared to Ada-QPacknet, which yields the best results in terms of capacity ($78.6\%$). The third strategy achieving satisfactory results is Packnet ($81.0\%$). 
The next strategies in the ranking are Cumulative and CWRStar, with a significantly lower accuracy ($36.52\%$ and $20.84\%$). 
In the 5 datasets scenario, Ada-QPacknet ($94.1\%$) presents a better performance than the second-ranked WSN ($93.41\%$) and the third-ranked Packnet ($92.59\%$) with a significant improvement for capacity ($33.7\%$) compared to Packnet ($82.86\%$) and WSN ($86.10\%$). We attribute this result to the low between-task similarity due to the different datasets, which makes the weight reutilization strategy adopted in Packnet and WSN less effective than Ada-QPacknet. The Cumulative approach is quite close to these forget-free methods with a performance of $84.44\%$. The other strategies achieve significantly worse results.  
In the TinyImagenet scenario, WSN achieves the best-performing score ($71.96\%$), slightly outperforming Ada-QPacknet  ($71.9\%$), which appears to be the second-best performing method. A possible explanation is that transfer learning via weight reutilization performed by other strategies is more effective in this scenario due to a higher task similarity when compared to 5 datasets. All the other strategies, such as Packnet ($55.46\%$), SI ($20.51\%$), Replay ($23.14\%$) are significantly worse and appear ineffective in this scenario.

The results presented in Table \ref{table:CL_capacity} show the comparison of the model's capacity in three different architectural CL strategies.
On p-MNIST, Packnet and WSN use $96.38\%$ and $77.73\%$ of the original model capacity, respectively. At the same time, Ada-QPacknet represents the middle ground, using $81.25\%$ of the capacity while achieving the best-performing accuracy scores.
In the case of s-CIFAR100, we also observe that forget-free strategies are the best performing. In this case, Ada-QPacknet (accuracy $74.1\%$, capacity $78.6\%$) represents the best approach in capacity when compared to  Packnet (accuracy $72.57\%$, capacity $81.0\%$) and WSN (accuracy $76.38\%$, capacity $99.13\%$), which suggests that the utilization of a higher capacity is a necessary condition to achieve higher accuracy on this dataset, given its complexity.
%
As for the 5-datasets scenario, Ada-QPacknet achieves better results than WSN and Packnet, while using a significantly lower model capacity ($33.7\%$ in comparison to over $80\%$ by both Packnet and WSN).
Finally, as for TinyImagenet, Ada-QPacknet presents second-best results in terms of capacity utilization ($112.5\%$), positioning itself in the middle between WSN ($48.5\%$) and Packnet ($188.67\%$), which appears particularly ineffective.
%

Overall, our results highlight that Ada-QPacknet achieves competitive results and compression ratios in all scenarios. It is noteworthy that Ada-QPacknet vastly outperforms other methods on the most complex 5-datasets scenario built upon heterogeneous datasets. 
One reason why Packnet and WSN manage to achieve similar results for p-MNIST and s-CIFAR100 could be their ability to share weights between tasks, while both aforementioned scenarios have homogeneous tasks originating from a single dataset. Therefore, weight sharing provides a huge advantage for WSN and Packnet.

\begin{table}
\small
\centering
\caption{Comparison with SOTA architectural and replay methods on Imagenet100 (10/10).}
\begin{tabular}{ccc} 
\hline
\textbf{Methods}   & \textbf{Parameters} & \textbf{Accuracy}  \\ 
\hline
\textbf{FOSTER}    &              -           &           74.49                       \\ 
\textbf{DyTox}        &   10.73M                       &      75.54                                        \\ 
\textbf{DER}        &      112.27M                    &          75.36                                     \\ 
\hline
\textbf{Ada-QPacknet}      &     11.5M                 &   72.26                      \\ 

\hline
\end{tabular}
\label{table:CL_capacity_new_competitors}
\end{table}

Additional results on the Imagenet100 scenario with hybrid methods are presented in Table \ref{table:CL_capacity_new_competitors}. The results highlight that Ada-QPacknet ($72.26\%$) is unable to outperform architectural methods extended with replay-based memory, such as FOSTER ($74.49\%$), DyTox ($75.54\%$), and DER ($75.36\%$). This phenomenon is expected due to the inherently higher complexity of hybrid approaches deriving from the memory component. Nevertheless, Ada-QPacknet performance appears still relatively close to hybrid methods, with a very low capacity occupation ($11.5$M), which is comparable to DyTox ($10.73$M) and significantly better than DER (112.27M) \footnote{For FOSTER, we do not report the number of parameters, since it  was not available in the original publication.}. 
We also note that these experiments were run without performing an extensive optimization for Ada-QPacknet hyperparameters (e.g., we used a fixed learning rate configuration) due to the extremely time-consuming requirements of the Imagenet100 scenario. For this reason, we expect that a better set of hyperparameters can be identified for our proposed method, which would position its accuracy closer to that of hybrid methods. 

\subsection{Ablation study}
We conduct experiments to verify that all stages of Ada-QPacknet contribute to the final model capabilities.
Our ablation study is twofold and aims at assessing: \textit{i)} model capacity (see Table \ref{table:ablation_capacity}), and \textit{ii)} model accuracy (see Table \ref{table:ablation_accuracy}) when a specific component of Ada-QPacknet is enabled.

Focusing on the most crucial factor - capacity, results in Table \ref{table:ablation_capacity} show that, on average, pruning allows to free a significant amount of model capacity ($45.0\%$ for p-MNIST, $50.0\%$ for s-CIFAR100, $52\%$ for 5-datasets). 
On the other hand, quantization frees, on average, a higher amount of capacity ($86.58\%$) for all three scenarios (p-MNIST, s-CIFAR100, and 5-datasets).
This behavior is expected since quantization has the potential to identify a low number of bits that retain most of the model accuracy, which is a finer-grained work (considering that a certain number of bits can be freed from each weight) when compared to adaptive pruning (which can only remove weight in its entirety). 
When combined, the two stages yield a remarkable capacity reduction per task ($91.88\%$ for p-MNIST, $92.30\%$ for s-CIFAR100, $90.10\%$ for 5-datasets).

Analyzing the results from the model accuracy viewpoint allows us to understand the trade-off between the model capacity (freed by adaptive pruning and quantization) and the corresponding model accuracy. 
Results in Table \ref{table:ablation_accuracy} show that quantization yields higher accuracy than pruning. It is clear that the removal of some weights may have more impact than reducing their bit-width in an adaptive and non-linear way. 
For example, for p-MNIST, the model with adaptive pruning yields an accuracy of $97.1\%$, and the quantization yields an accuracy of $98.5\%$. The other datasets follow a similar pattern with quantization yielding $75.8\%$ and $95.2\%$ accuracy for s-CIFAR100 and 5-datasets respectively, where pruning yields an accuracy of $75.3\%$ and $94.4\%$. 
As for the final Ada-QPacknet, the accuracy is equal or lower than that of a single component: $97.1\%$, matching the performance of pruning, which is lower than $98.5\%$ (quantization) for p-MNIST, whereas it drops from $75.3\%$ (pruning) and $75.8\%$ (quantization) to $74.1\%$ for s-CIFAR100, from $94.4\%$ (pruning) and $05.2\%$ (quantization) to $94.1\%$ for 5-datasets. However, a degree of reduction is expected since the model is being simplified by adaptive pruning and quantization, and it is reassuring to see that the performance drop is not significant.

Overall, we conclude that both adaptive pruning and quantization contribute to model capacity reduction,  enabling the model to accommodate future tasks without significantly impacting model accuracy.
We note that results for our method without capacity reduction pruning and quantization are missing, as their interpretation would be cumbersome. Conceptually, Ada-QPacknet without pruning and quantization could be reformulated in two ways: \textit{i)} behave similarly to Naive, where the same model is fully retrained as a new task is observed, leading to forgetting; \textit{ii)} being a forget-free model, which, however, would fully saturate its capacity after the first task, invalidating its lifelong learning capabilities.

\begin{table}
\small
\centering
\caption{Capacity of each component and both combined (Ada-QPacknet), average per each task}
\label{statistc_of_datasets}
\begin{tabular}{ccccc} 
\hline
\textbf{Datasets}   & \textbf{p-MNIST} & \textbf{s-CIFAR100} & \textbf{5 datasets}  \\ 
\hline
\textbf{pruning}    &        45.0\%                  &          50.0\%      &       52.0\% \\
\textbf{quantization}        &      12.5\%                     &      15.25\%                &             12.5\%    \\
\hline
\textbf{Ada-QPacknet}      &     8.12\%               &   7.7\%             &   9.9\%          \\ 
\hline
\end{tabular}
\label{table:ablation_capacity}
\end{table}

\begin{table}
\small
\centering
\caption{Accuracy of each component and both combined (Ada-QPacknet)}
\label{table:ablation_accuracy}
\begin{tabular}{ccccc} 
\hline
\textbf{Datasets}   & \textbf{p-MNIST} & \textbf{s-CIFAR100} & \textbf{5 datasets} \\ 
\hline
\textbf{pruning}    &         97.1                 &   75.3               &     94.4                           \\ 
\textbf{quantization}        &       98.5                    &         75.8              &       95.2           &                \\ 
\hline
\textbf{Ada-QPacknet} &          97.1             &        74.1              &       94.1    & \\
\hline
\end{tabular}
\label{accuracy}
\end{table}


In addition to our ablation study, we report some patterns that emerged during our experiments. For most of the scenarios tasks, the performance drop observed with 4-bit quantization was less than $1\%$ for all tasks (in comparison to full 32-bit weights). For example, in p-MNIST, a performance drop of $-0.5\%$, $-1.9\%$,  and $-3.0\%$ was observed with 3, and 2 bits, respectively. 
In 5 datasets, tasks have shown a different degree of complexity ranging from low (3 bits were used for p-MNIST, notMNIST, and s-CIFAR100), medium (4 bits for Fashion MNIST), and high (5 bits for SVHN). For Imagenet100, the least configuration yielding a negligible drop in accuracy was 5 bits.
%
It should be noted that for TinyImagenet, the overall capacity was decreased by allocating 5 separate subnetworks, each with 8 tasks. This choice allowed us to minimize the number of bits per mask, which would otherwise be 40 bits due to the 40 tasks included in this dataset.

From a computational complexity viewpoint, we observed that the training time in Ada-QPacknet is dominated by lottery ticket search, whereas quantization adds a negligible cost to the execution time since it is done post-training.
Comparing our method with WSN and PackNet, we observed an overhead of about $50\%$ in terms of overall execution time.
This result shows that the positive results obtained in terms of model accuracy and capacity for Ada-QPacknet are achieved with a reasonable execution time that fundamentally lies in the same complexity class as its direct competitors.

\section{Conclusions and future works}
\label{sec:conclusions}
In this paper, we proposed Ada-QPacknet, a novel forget-free architectural method for continual learning that resorts to adaptive pruning and adaptive quantization. 
The adaptive pruning approach allows the method to identify the near optimal subset of weights that preserves the performance of the model on single tasks while minimizing the capacity required to solve each task.
Moreover, the adaptive quantization strategy adopted in our method allows the use of a single weight for more than one task without any significant drop in performance. 
The presented results show that our proposed method achieves competitive results in terms of accuracy and utilization of model capacity than many continual learning strategies, including popular forget-free methods. 
In future work, we will focus on devising more robust modeling capabilities to better deal with complex scenarios (such as Imagenet100) where Ada-QPacknet achieved a suboptimal performance. Specifically, we will focus on estimating layer sensitivities and correlations between task distributions in order to estimate sparsity-level boundaries for new tasks. 
Moreover, 
we will investigate the adoption of weight sharing, to achieve higher compression ratios per task, as well as hybrid quantization.
Finally, we will devise a mechanism to select the most suitable mask among the available ones allocated to previously learned tasks once full model capacity is reached.

\section*{Acknowledgments}
This research was supported in part by the PLGrid Infrastructure.

\bibliographystyle{unsrt}

\end{document}